\documentclass[twoside,11pt]{article}
\usepackage{jair,rawfonts}
\usepackage{apacite}
\jairheading{1}{2021}{1-15}{6/21}{/21}
\ShortHeadings{Image Captioning As an Assistive Technology}
{Dognin et al }
\firstpageno{1}
\newcommand*{\escape}[1]{\texttt{\textbackslash#1}}
\usepackage{amsmath}
\usepackage{color}
\usepackage{multirow}
\usepackage{url}

\usepackage{microtype}
\usepackage{booktabs} 
\usepackage{graphicx}
 \usepackage[caption=false,font=footnotesize]{subfig}
\usepackage[thinlines]{easytable} 
\usepackage{bm}
\usepackage{xcolor}
\usepackage{hhline}
\begin{document}





\title{Image Captioning as an Assistive Technology: Lessons Learned from VizWiz 2020 Challenge  }
%
%
%
%
\author{\name Pierre Dognin$^*$ \email pdognin@us.ibm.com \\
\name Igor Melnyk$^*$ \email igor.melnyk@ibm.com\\
\name Youssef Mroueh$^{*}$ \email mroueh@us.ibm.com\\
\name  Inkit Padhi$^*$  \email inkit.padhi@ibm.com \\ 
\name  Mattia Rigotti$^*$ \email mrg@zurich@ibm.com \\
\name Jarret Ross$^*$ \email rossja@us.ibm.com \\
\name  Yair Schiff$^*$  \email yair.Schiff@ibm.com  \\
\name Richard A. Young \email richard.young2@ibm.com \\
\name Brian Belgodere \email bmbelgod@us.ibm.com\\
\addr IBM Research AI \thanks{ Authors contributed equally, alphabetical order   } }

\maketitle
\begin{abstract}
Image captioning has recently demonstrated impressive progress largely owing to the introduction of neural network algorithms trained on curated dataset like MS-COCO.
Often work in this field is motivated by the promise of deployment of captioning systems in practical applications.
However, the scarcity of data and contexts in many competition datasets renders the utility of systems trained on these datasets limited as an assistive technology in real-world settings, such as helping visually impaired people navigate and accomplish everyday tasks.
This gap motivated the introduction of the novel VizWiz dataset, which consists of images taken by the visually impaired and captions that have useful, task-oriented information.
In an attempt to help the machine learning computer vision field realize its promise of producing technologies that have positive social impact, the curators of the VizWiz dataset host several competitions, including one for image captioning.
This work details the theory and engineering from our winning submission to the 2020 captioning competition.
Our work provides a step towards improved assistive image captioning systems.
\end{abstract}




%

\section{Introduction}\label{sec:introduction}


%
%
%
%


 \begin{figure*}[ht!]
 \centering
 	\includegraphics[width=6.1in,keepaspectratio]{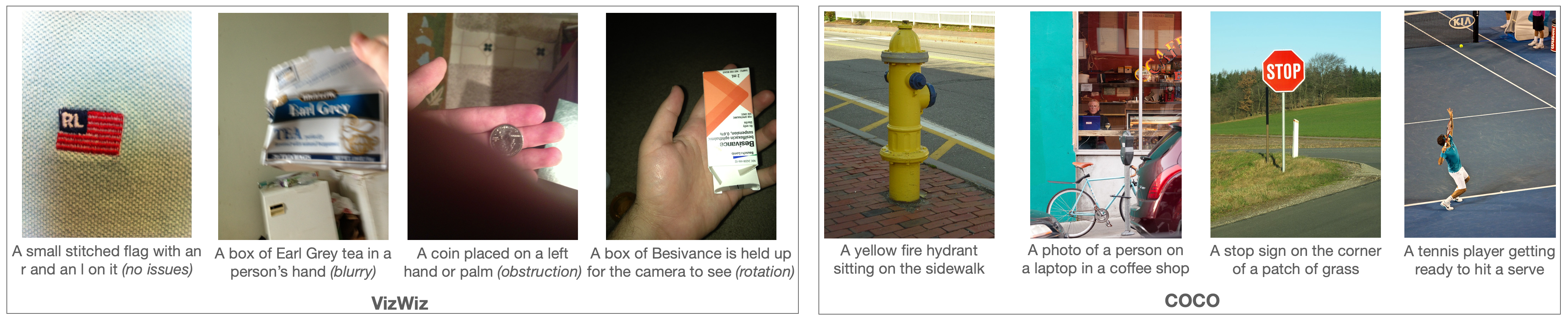}
 	\label{fig:dataset_comparison}
 	\caption{Comparison of images in VizWiz (goal or task oriented captions) and COCO dataset (generic captions). For VizWiz images, along with the caption, we note any quality issues that are present in the image, such as blur, need for rotation, and occlusions.}
 	\label{fig:vizwizVSCoco}
 \end{figure*}
 
Image captioning has witnessed steady progress since 2015, thanks to the introduction of neural caption generators with convolutional and recurrent neural networks \cite{GoogleNIC,Karpathy}.
Such progress, however, has been by and large demonstrated on curated datasets like MS-COCO\cite{MSCOCO}, whose limited size and scarcity of contexts result in image captioning systems that tend to produce terse and \emph{generic descriptive} captions.
This drastically restricts the applicability of image captioning algorithms and prevents them from being widely deployed in practical real-world settings, such as helping visually impaired people navigate and accomplish everyday tasks.

The realization of these limitations and the drive to produce image captioning systems that constitute mature assistive technologies motivated the introduction of the VizWiz Challenges for captioning images taken by people who are blind in \cite{Gurari_2018_CVPR}, \cite{vizwiz_challenge} and\cite{cap_vizwiz}.
The VizWiz Challenges are the result of a paradigm shift of image captioning towards \emph{goal-oriented captions}, i.e.\ captions that not only faithfully describe a scene from everyday life, but also answer specific needs, such as those that might help blind users achieve particular tasks.
A recent concurrent work to ours, refers to goal-oriented captioning as captioning with purpose \cite{fisch_capwap_2020}. Figure \ref{fig:vizwizVSCoco} contrasts the difference between goal-oriented (VizWiz) and generic image captioning (COCO); for example a goal-oriented caption helps the visually impaired in finding the label of a medicine box as compared to a generic description that does not specify the type of box and its label. The VizWiz 2020 Image captioning \cite{cap_vizwiz} challenge was designed to address challenges in accessibility and to foster advances in designing and building assistive captioning systems from images taken by the visually impaired.

In this paper, we describe our winning entry to the VizWiz 2020 Image Captioning Challenge, and highlight the lessons we learned in terms of system design from both \emph{\textbf{accessibility}} and \emph{\textbf{utility}} perspectives.

Firstly on accessibility, images taken by visually impaired people are captured using phones and may be blurry and have flipped orientations.
The paper \cite{Chiu_2020_CVPR} gives a comprehensive account of the quality challenges encountered in the real world images taken by the blind, such as blur, poor lighting conditions, non-centered framing, occlusion, and rotation issues. Examples of these image quality challenges can be seen in Figure \ref{fig:vizwizVSCoco}.
Ensuring that our machine learning pipeline is \emph{robust} to these image quality issues and can compensate for rotation angles of the image framing is a crucial part of our work.
In Section \ref{sec:ImFeature}, we show how we address some of these challenges via the use of a ResNeXt network \cite{ResNext} that was pretrained on billions of Instagram images taken using phones. In Section \ref{subsubsec:ocr}, we tackle the rotation issue by building a dictionary-guided rotation-invariant text recognizer that registers images to their canonical orientation and enables rotation invariant text recognition.
 
 Secondly on utility, we augment our system with reading (Sections \ref{subsubsec:ocr} and \ref{sec:InvOCR}) and semantic scene understanding capabilities (Section \ref{sec:OBJ}). Many of the VizWiz images have text that is crucial to the goal and the task at hand. We equip our pipeline with optical character detection and recognition, OCR (\cite{baek2019character} and \cite{ baek2019STRcomparisons}).
 Then, to ensure the robustness of OCR to rotations as discussed earlier, we perform OCR on four orientations of the image and select the orientation that has a majority of sensible words in a dictionary. In order to improve the semantic understanding of the visual scene, in addition to ResNeXt image features (IMG) and OCR, we augment our pipeline with object detection and recognition pipeline \cite{tan2020efficientdet} (OBJ) that are discussed in Section \ref{sec:OBJ}.
 
 As we describe in Section \ref{sec:models}, our model architecture then fuses visual features with detected texts and objects that are embedded using fastText \cite{bojanowski2016enriching}  with a multimodal transformer\cite{Transformer}. To ensure that vocabulary words coming from OCR and object detection can be used for out-of-vocabulary generation, we incorporate a copy mechanism \cite{singh_towards_2019,yao2017copy,gu2016copy} in the transformer that allows it to choose between copying an out-of-vocabulary token or predicting an in-vocabulary token. We train our system using Cross-Entropy (CE) pre-training and fine-tune using Self-Critical Sequence Training (SCST) \cite{scst} optimizing CIDEr \cite{CIDEr} score.
 
 Finally, we give experimental details and extensive ablations studies on the VizWiz challenge and the competition results in Section \ref{sec:experiments}. In Section \ref{sec:qualitative}, we discuss qualitative evaluation of our winning entry and how it compares to models trained on generic image captions such as MS-COCO \cite{MSCOCO} or Google Conceptual Captions (GCC) \cite{sharma2018conceptual}. In Section \ref{sec:demo}, we describe a real-time demo of our system that uses our captioning pipeline and the Watson Text to Speech API \cite{watsonTS}, showing the  potential of image captioning as an assisting technology for the visually impaired.

\section{Related Work}

\textbf{Generic image captioning.} Recent progress in captioning since the seminal Neural Baby Talk work of \cite{baby_talk} in the pre-deep learning era has been accelerated by the wider adoption of deep sequence-to-sequence models. Neural Image Captioning (NIC) on MS-COCO \cite{MSCOCO} introduced the influential work of Show-and-Tell using LSTM architectures and cross-entropy training \cite{Karpathy,GoogleNIC}. NIC won the MS-COCO 2015 challenge \cite{Vinyals_2017}.
Attention models introduced in \cite{CapAttention,xu2015show} further improved performance in image captioning and were refined in bottom-up and top-down attention models \cite{Anderson2017up-down}.
Transformers models \cite{Transformer} have been adapted to multimodal scenarios, such as image captioning and visual question answering (VQA) in works like \cite{kant2020spatially} and  \cite{googlewinner}, which won the conceptual captions challenge on GCC dataset in 2019 \cite{sharma2018conceptual}. Generic image captioning systems were trained on MS-COCO or GCC benchmark using cross-entropy training. Improvements were further obtained using automatic metric training, such as CIDEr and SPICE, combined with Reinforcement Learning techniques, as in \cite{Ranzato,scst,spider}. 
Generation of more natural sounding captions was also addressed by the use of GAN losses in \cite{bodai,shetty2017speaking,dognin2019adversarial} and better sequence-to-sequence architectures such as Transformers \cite{Transformer} as in \cite{googlewinner}.  
Copy mechanisms in image captioning and VQA ensure generalization to words not in the vocabulary that are available from other channels, such as object or text detectors \cite{singh_towards_2019,yao2017copy,gu2016copy}.

\noindent\textbf{Accessibility and assistive technologies challenges and needed functionalities.} VizWiz Social is a mobile application centered around accessibility whereby visually impaired users can take pictures, ask questions about them and crowd-source real-time answers \cite{vizwiz_app}.
\cite{brady_visual_2013} analyzed the type of images uploaded and questions asked by the users on this app and grouped them into the following categories: identification, reading, and description \cite{brady_visual_2013}. \cite{kacorri2017people} studied the training of personalized object recognition using VizWiz social. Assistive technologies and machine learning based systems require a set of functionalities to meet the accessibility needs of the visually impaired such as: object, text, and color recognition, as well as counting \cite{zeng_vision_2020}. 

\noindent \textbf{VizWiz challenges and datasets.} Using the data collected from the VizWiz Social application, several VizWiz datasets and challenges have been proposed on VQA in 2018 \cite{Gurari_2018_CVPR, vizwiz_challenge} and the image captioning challenge in 2020 \cite{cap_vizwiz}. The latter is the main focus of our work.

\noindent{\textbf{Assistive Technologies}} While the space of assistive captioning technology remains an open domain, there are certain solutions available that have begun to address this important need using a variety of mobile applications, wearable devices, and other assistive hardware.
For example, ORCAM \cite{ORCAM} is a wearable device solution that combines voice-activated commands with text, object, and facial detection to help users read text and recognize faces and products on-demand.
A different approach comes in the form of the NavCog \cite{NavCog} application and smart guiding suitcase prototype, which utilizes radio and Bluetooth signals combined with machine learning optimization to determine a user's location and assist the visually impaired in navigating to a desired destination.
Finally, the solution that most closely aligns to the goal-oriented automated captioning, addressed in this work, is found in the Seeing AI mobile application \cite{MSR_seeingAI}, which provides text reading, currency, product, and facial recognition, and scene description capabilities.

\vskip-0.2in
\section{Model}\label{sec:model}

\begin{figure*}[!t]
\centering
\includegraphics[width=\linewidth]{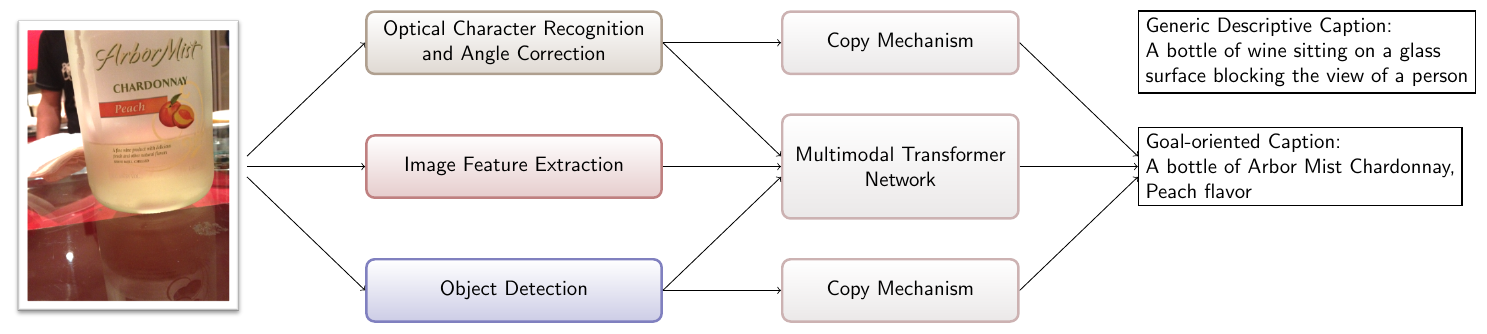}
\caption{Multimodal Assistive Captioner Overview. An input image is passed through 3 separate modules to perform image feature extraction, OCR and angle correction, and object detection. Their outputs are combined and submitted to a Multimodal Transformer Network. The generated caption is further biased towards the OCR and object detection outputs by means of copy mechanisms to produce more goal-oriented content, rather than a generic descriptive caption.}
\label{fig_cap}
\end{figure*}

\begin{figure*}[!t]
\centering
\includegraphics[width=\linewidth]{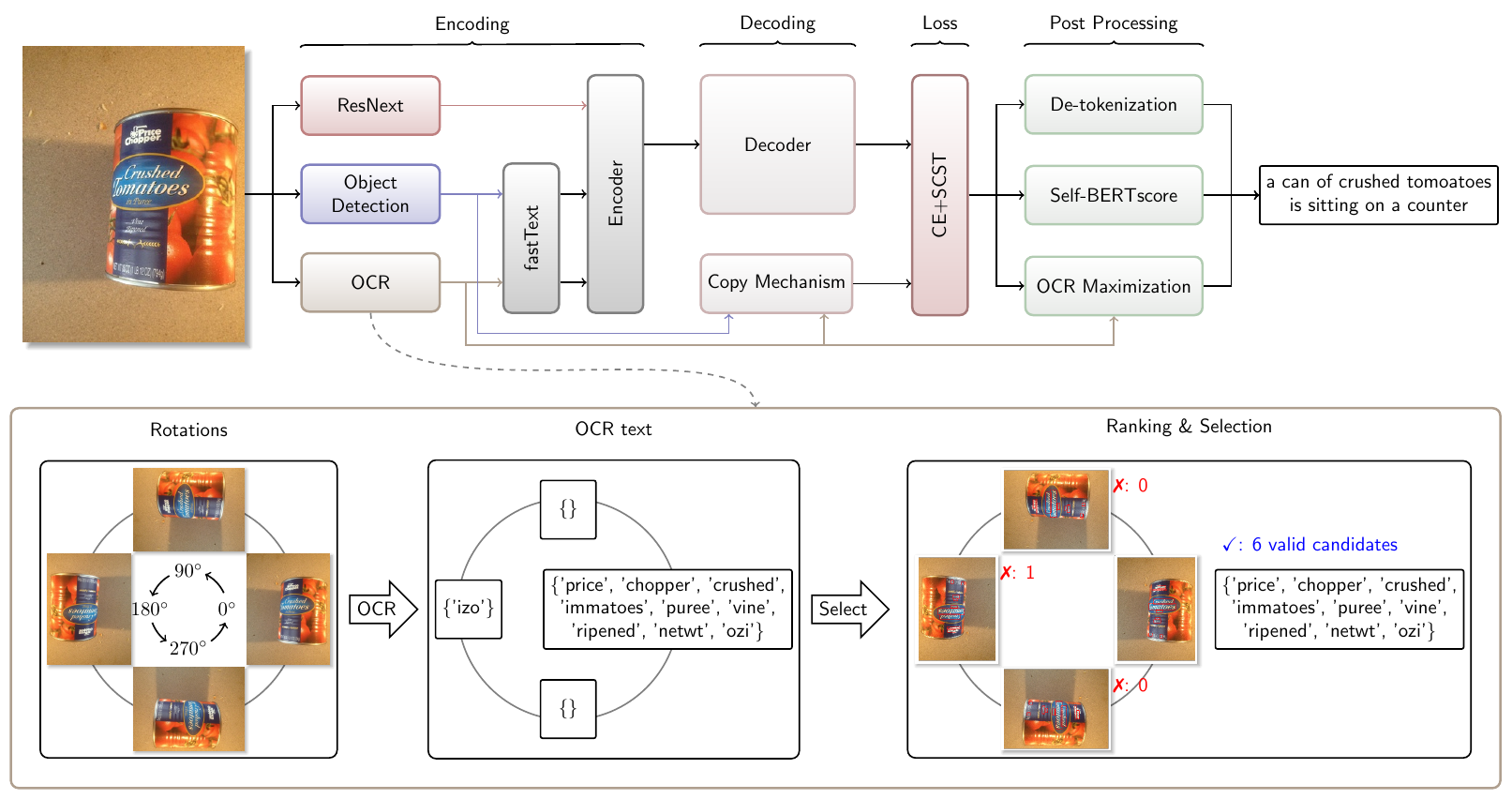}
\caption{Multimodal Assistive Captioner. Our captioner is a multimodal Transformer with an encoder/decoder architecture composed of 4 distinct stages. First, an Encoding stage provides embeddings for the 3 modality streams of image features, object detection, and OCR results derived from an input image. Image features are extracted by a ResNeXt model, while object detection and OCR modules generate sequences of object categories (10 max) and strings of characters (20 max) respectively, which are then embedded using fastText. Embeddings for these 3 modalities are combined and fed to a Decoding stage where a decoder generates a caption whose content is biased towards OCR and object detection results by a copy mechanism that influences the Loss computation stage. The Post Processing stage finalizes the generated caption by cleaning up the caption, maximizing the use of OCR tokens, and promoting coherent candidate (via self-BERT score). The OCR module is expanded to show the inner workings of our Dictionary-Guided Rotation-Invariant OCR module. The input image rotations lead to 4 sets of OCR results which enable the selection of a proper rotation based on the largest number of intelligible tokens.}
\label{fig_mmcap}
\end{figure*}

In this section, we describe our model and the different data channels we incorporate to address task-oriented image captioning.  
An overview of our multimodal assistive captioner is given in Figure~\ref{fig_cap}. As seen there, in addition to the image feature extraction discussed in Section \ref{sec:ImFeature}, our system incorporates OCR and object detectors that are detailed in Section \ref{sec:OCRandOBJ}. A technical overview  is given in Figure \ref{fig_mmcap}.

\subsection{Image Feature Extraction with ResNeXt}\label{sec:ImFeature}

For image processing, we choose ResNeXt \cite{ResNext} as our feature extractor since we observed a 10-point CIDEr gain compared to ResNet 101 features \cite{he15deepresidual} (all else equal). We hypothesize that this gain comes from the fact that VizWiz images, which are mainly captured using cell phone cameras, more closely align to the ResNeXt pre-training that uses Instagram phone-taken images.    
We preserve the size of the image as long as one of the sides is greater than 320 pixels. 
If this requirement is not met, we increase the offending image axis to 320 pixels while maintaining the aspect ratio. We normalize images using the method from \cite{ResNext}. To extract features, we use the 99th layer of the 101 layer ResNeXt \cite{ResNext} convolutional neural network with cardinality of 32 and bottleneck dimension of 8. 
These features are finally pushed through a 2-dimensional adaptive max pooling layer forming a 14$\times$14$\times$2048 feature vector for each image.
Before being input into our multimodal Transformer, the image feature vector is reshaped to 196$\times$2048.
\subsection{Incorporating OCR and Objects}\label{sec:OCRandOBJ}

\subsubsection{Incorporating OCR}\label{subsubsec:ocr}


\begin{figure*}
\vskip -0.6in
\centering
\includegraphics[width=\linewidth]{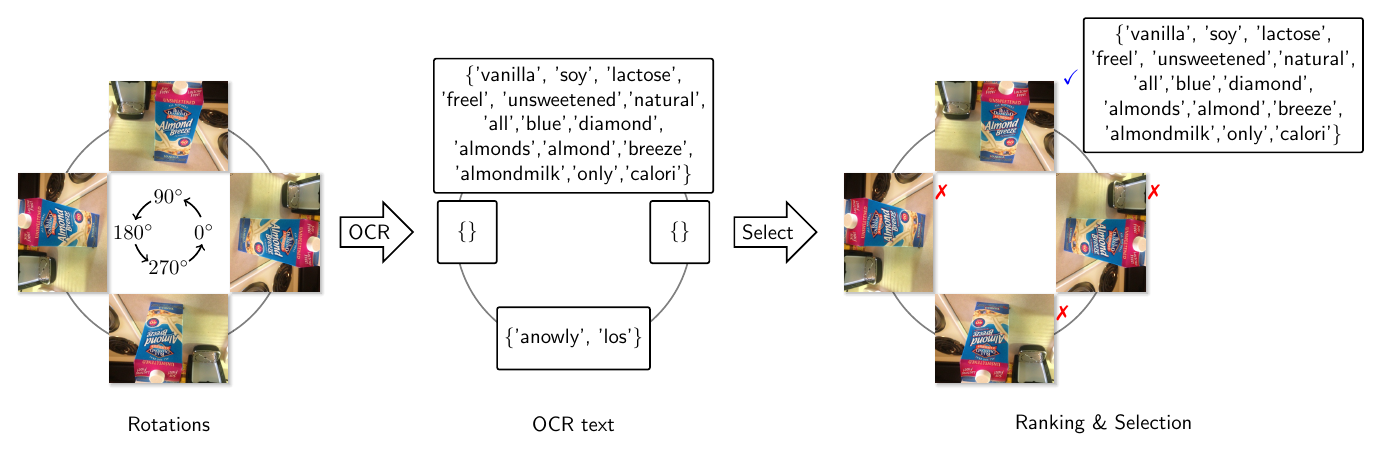}
\caption{OCR example with two competing rotations. In this case, OCR detects tokens for two rotations, and after dictionary-guided ranking the tokens associated with $90^{\circ}$ rotation are selected.}
\label{fig_ocr2}
\vskip -0.05in
\end{figure*}
\begin{figure*}[ht!]
\vskip -0.3in
\centering
\includegraphics[width=\linewidth]{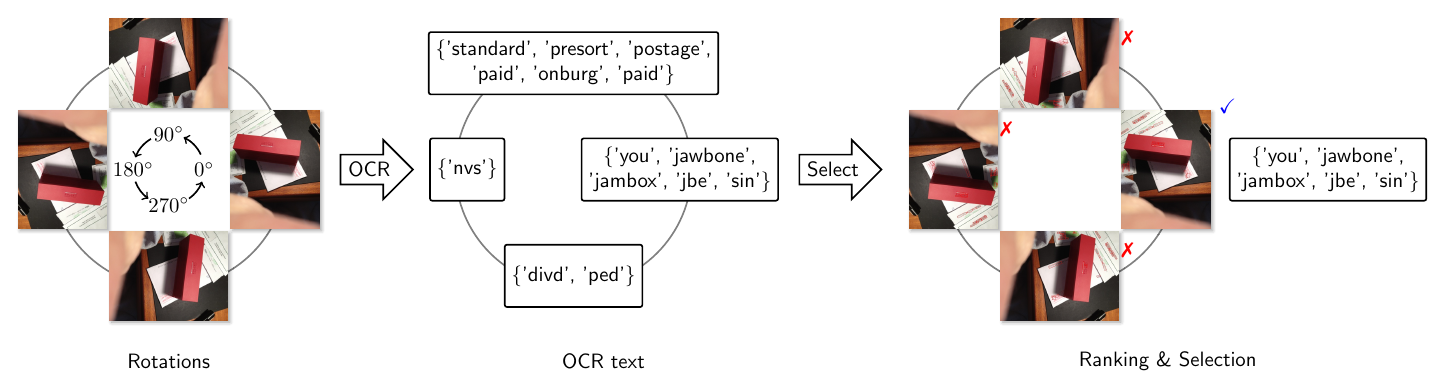}
\caption{OCR example with four competing rotations. In this case, OCR detects tokens for all four rotations, and after dictionary-guided ranking the tokens associated with $0^{\circ}$ rotation are selected.}
\label{fig_ocr4}
\end{figure*}

Manual inspection of the VizWiz images and captions reveals that OCR is an integral component of building meaningful captions for images that are provided by visually impaired users.
Product labels, nutrition facts, instructions, and other relevant text on objects and screens are often found in the VizWiz dataset images.
Therefore, when composing our pipeline to tackle this challenge, it was evident that a state of the art OCR component would be critical to render our captioning system useful.

To that end, we investigated several open source OCR solutions, and eventually decided to incorporate \cite{baek2019character} and \cite{ baek2019STRcomparisons}.
Having demonstrated success reading text in various fonts and lighting conditions, this open source solution was a useful addition to our pipeline.

\subsubsection{Dictionary-Guided Rotation-Invariant OCR}\label{sec:InvOCR}
Although this open source implementation provides strong OCR capabilities, we noted that it did contain some failure modes, specifically with regards to text orientation.
The nature of the VizWiz dataset made this a unique obstacle for this accessibility challenge, as images often have non-horizontal orientations: approximately 28.5\% of the training images and 28.3\% of validation images have rotation issues, according to the metadata annotations on the competition website. 
To address this shortcoming in the OCR module, we therefore augment the OCR component of our pipeline by pre-processing images in four possible orientations, i.e. 0, 90, 180, and 270 degree rotations.
Specifically, for each rotation, we pass it through the OCR module and extract the predicted text.
The OCR predictions are also passed to the fastText \cite{bojanowski2016enriching} tokenizer.
The orientation that produces the greatest number of intelligible tokens, according to the fastText vocabulary, is selected.
We also observed that in images which contain a lot of text, the detected texts that contribute to the caption the most are usually the ones having the largest bounding box.
This inspired us to rank detected OCR texts by bounding area and text confidence in descending order.

In Figures \ref{fig_ocr2} and \ref{fig_ocr4}, we present schematic representations of the OCR modality for two different images. For each image, we consider four rotations which are then passed through the OCR module. These examples highlight how the OCR module is sensitive to rotations and, in some cases, yields no legible text. Through the inclusion of dictionary-guided ranking, we select OCR text(s) from a particular view, making the overall modality invariant to the rotation. The final step of the OCR module involves creating features using fastText embeddings of the final text(s).


\subsubsection{Object Detection}\label{sec:OBJ}
In addition to the OCR component, we also included an object detection module (OBJ) to aid the captioner in composing more detailed and accurate image descriptions. After reviewing several popular off-the-shelf solutions, we decided to use EfficientDet from \cite{tan2020efficientdet}. The main trade-off in selecting the detector is between the number of recognized objects versus the detection accuracy. Although EfficientDet covers only 80 objects based on MS-COCO categories \cite{MSCOCO}, its accuracy is high enough to be a reliable input source. This is in contrast to other approaches (e.g, YOLO9000 \cite{redmon2016yolo9000}), which offer larger object coverage but suffer from less reliable detection accuracy.  
For the VizWiz challenge, the EfficientDet was used as is, without any modifications or adaptations.
Given an image, the OBJ detector returns a list of objects and the corresponding confidence levels.
We then sort the objects based on the confidence and select the ones above $0.25$.
Similar to the OCR module, the detected object strings are encoded using fastText and passed to the captioner as a separate modality. 
 

\section{Multimodal Transformer}\label{sec:models}

\subsection{Regular Transformer (without copy mechanism)}
\begin{figure}
\centering
\includegraphics[width=0.5\linewidth]{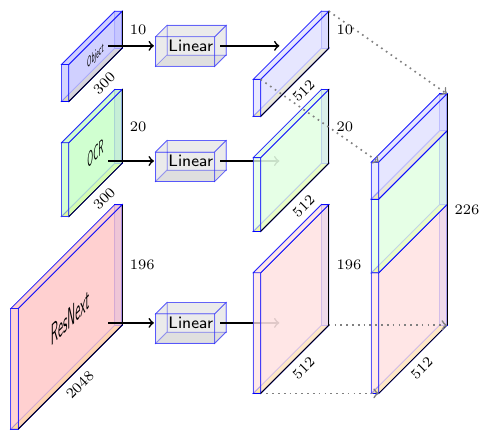}
\caption{Multimodal features transformation and concatenation.}
\label{fig_concat}
\end{figure}

Given the image features extracted from ResNeXt that we reshape to a sequence of $n_{\text{pixel}}\times$2048 (where $n_{\text{pixel}}=196$), detected objects (clipped to a maximum $n_{\text{obj}}=10$ objects), and OCR (clipped to maximum $n_{\text{ocr}}=20$ words) embedded in fastText to 300-dimensional vectors each, we embed those three modalities to a joint space of dimension $d=512$ using linear projections. Then, we concatenate the three modalities resulting in a sequence of length $n=n_{\text{pixel}}+ n_{{\text{ocr}}}+n_{\text{obj}}$ of dimension $d$. This simple concatenation is then fed to a regular transformer \cite{Transformer}.
Note that the self-attention in this case is a great model for intra and inter correlations between the modalities: the diagonal of the self-attention matrix represents self-attention on each modality alone, and the off-diagonal terms represent cross attention between modalities (IMG, OCR), (IMG, OBJ), and (OBJ, OCR).
A summary of this step is given in Figure~\ref{fig_concat}.

\begin{figure}[!t]
\centering
\includegraphics[width=0.5\linewidth]{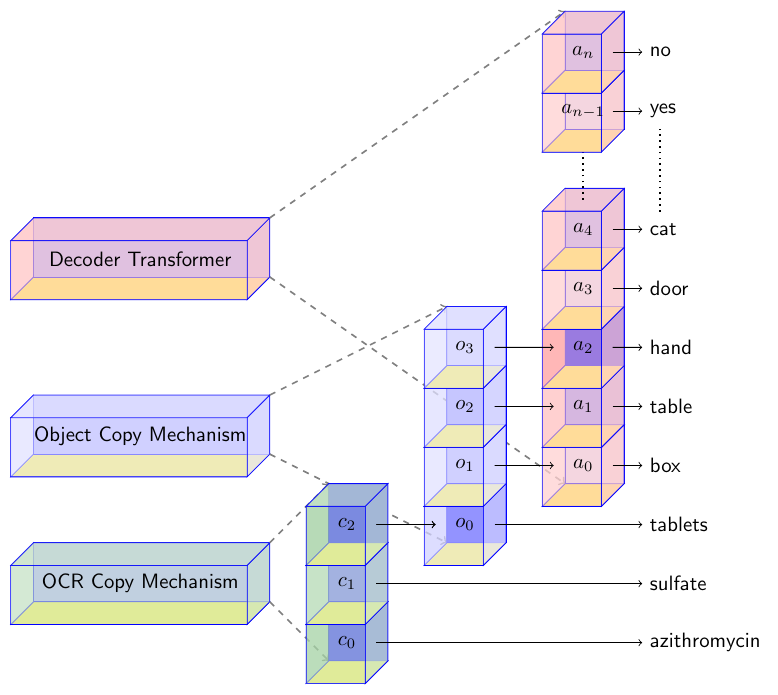}
\caption{Biasing softmax and vocab expansion with copy mechanism for object detection and OCR.}
\label{fig_copy_mech}
\end{figure}

\subsection{Copy Transformer and Dynamic Vocabulary}
\label{sec:copy}
OCR and OBJ are important input modalities for building semantically meaningful representations of a scene.
In addition, segments from the OCR and OBJ can be directly useful as parts of the output sequence.
Specifically, giving the decoder transformer the option to sample from the OCR and OBJ outputs is a straightforward way to extend the vocabulary of the captions with \emph{out-of-vocabulary tokens} that are potentially relevant to describing the scene.
A mechanism that is naturally suited to achieve this is the ``copy mechanism'', which was originally proposed for seq2seq NLP tasks in \cite{gu2016copy} as a way to accommodate both understanding of the input sequences and rote memorization of parts of it and direct copy in the output sequence.
\cite{yao2017copy} subsequently incorporated copying in image captioning, and \cite{see2017pointer} use a similar mechanism for summarizing text.

Inspired by this line of work, we decided to implement a copy mechanism that optionally copies tokens generated by the OCR and OBJ modules directly to the output.
This is obtained by expanding the vocabulary of the decoder transformer with the outputs of the OCR and OBJ modules.
Concretely, we do this by concatenating the caption and OCR and OBJ vocabularies. We then mask out at caption generation the unnormalized logits associated with tokens that were not present in the OCR and OBJ outputs.
We compute the probability of sampling each one of the tokens in the resulting expanded vocabulary by applying a softmax non-linearity on the vector of concatenated unmasked unnormalized logits.
As a result, the probability that the decoder transformer samples a token $y_t$ at generation step $t$ conditioned on an image and the previously generated tokens can be written as a normalized average of probabilities:
\begin{align*}
    p(y_t|\cdot) = \frac{1}{Z}\left(e^{\phi_{\text{IMG}}(y_t)}+e^{\phi_{\text{OCR}}(y_t)}+e^{\phi_{\text{OBJ}}(y_t)}\right),
\end{align*}
where $Z$ is a normalization term, and $\phi_{\text{IMG}}(\cdot)$, $\phi_{\text{OCR}}(\cdot)$, $\phi_{\text{OBJ}}(\cdot)$ are score functions outputting unnormalized logits over the captions, OCR, and OBJ vocabulary, respectively, with the understanding that $\phi_{\text{OCR}}(\cdot)$ is set to $-\infty$ for tokens that do not appear in the output of the OCR module, and $\phi_{\text{OBJ}}(\cdot)$ is set to $-\infty$ for tokens that do not appear in the output of the OBJ module.
In practice, this corresponds to biasing the caption generator to sample from the outputs of the OCR and OBJ modules, as illustrated in Figure~\ref{fig_copy_mech}, and most importantly, it endows our captioning architecture with a ``dynamic vocabulary'' that comprises out-of-vocabulary tokens that might not be present in the caption vocabulary, but are captured by the OCR and OBJ modules.

\section{Experiments}\label{sec:experiments}

\subsection{Dataset}
We use the VizWiz Captions dataset for all our experiments.
This dataset is split into training, validation, and testing subsets as described in Table~\ref{tbl:dataset}.
\begin{table}
\centering
\caption{VizWiz Captions dataset information.}
\begin{tabular}{lrr}
\toprule
Subset & Images & Captions \\
\midrule
Training & 23,431 & 117,155 \\
Validation & 7,750 & 138,750 \\
Testing & 8,000 & 40,000 \\
\bottomrule
\end{tabular}%
\label{tbl:dataset}
\end{table}
Given the original dataset, we prepare the data for training by first removing any special characters, such as \escape{n} (new line), \escape{r} (carriage return), and \escape{t} (tab), from the captions.
Then, we remove any punctuation characters from the captions before taking their lowercase versions.
Finally, each caption is tokenized using the BERT tokenizer, with any words not in the BERT dictionary replaced by the \texttt{<unk>} token.
At training time, the captions, together with OCR and OBJ detected tokens bring the total of observed unique tokens to 35,555.

\subsection{Training Details}\label{sec:details}
In this section, we discuss the details of our model training process.
We start with the basic Cross-Entropy (CE) pre-training, then we randomly select one CE model (usually close to the end of the training) to initialize the Self-Critical Sequence Training (SCST) step.
We repeat the above CE+SCST steps to build a multitude of models by varying the model hyper-parameter values (such as number of Transformer layers, hidden dimension size, random seed initialization, etc), varying the set of used data modalities (IMG, OCR and OBJ), and turning the copy mechanism on/off.
Finally, we combine these model permutations into an ensemble for evaluation and testing.\\

\noindent \textbf{CE Training} The CE training is run for 10 epochs, using a batch size of 80.
We employ SGD with ADAM optimizer with $(\beta_1,\beta_2)=(0.9, 0.98)$.
We warm the learning rate with a factor of 1 for 2000 iterations (minibatches) and then decay it  proportionally to $1/\sqrt{i}$, where $i$ is the iteration step. \\

\noindent \textbf{Evaluation Metrics} To examine the quality of the generated image descriptions, several automatic metrics for caption evaluations have been proposed. We briefly review a few of them that will be used in this paper. 

\noindent\textit{BLEU} (BiLingual Evaluation Understudy) \cite{papineni2002bleu} is a popular and widely used metric that focuses on precision.
Its value is computed as the geometric mean of
n-gram (in this work we use 4-gram) matching between the generated caption and multiple references with a brevity penalty to discourage shorter captions. 

\noindent \textit{METEOR} (Metric for Evaluation of Translation with Explicit ORdering) \cite{banerjee2005meteor} is a metric that is based on the harmonic mean of unigram precision and recall, and is computed via an explicit word-to-word matching (or synonym matching) between generated and one or more reference sentences.

\noindent \textit{ROUGE} (Recall-Oriented Understudy for Gisting Evaluation) \cite{lin-2004-rouge} is a recall-based metric and thus has a tendency to
reward long detailed sentences. The value is usually computed based on n-gram recall between the candidate and the references. Alternatively, longest common sub-sequence-based statistics is also frequently applied, as in this work. 

\noindent \textit{CIDEr} (Consensus-based Image Description Evaluation) \cite{vedantam2015cider} is a popular metric that uses TF-IDF scores (to decrease the influence of frequent, less informative words) to weight shorter and longer n-grams across the reference captions, where the higher order n-grams can usually better capture the grammatical properties as well as encode richer semantics.

\noindent \textit{SPICE} (Semantic Propositional Image Caption
Evaluation) \cite{anderson2016spice} is another metric which, in contrast to the above examples, does not use n-grams but rather estimates the caption quality by transforming the candidate and references into a semantic representation called a scene graph, that is eventually converted into a set of word tuples.
The final metric is computed as an F-score over these tuples.\\

\noindent \textbf{SCST Training} Recall, that in CE training we maximize the likelihood of the next ground-truth token given the previous ground-truth token, while at test time the next token is predicted given previously self-generated tokens.
This creates a train-test mismatch. Moreover, although CE loss is used for training, the evaluation metrics such as BLEU, CIDEr, etc. are used for testing, again creating a mismatch.
To address these issues, we employ Self-Critical Sequence Training (SCST) to fine-tune our model after the CE training.
SCST is a variant of the REINFORCE algorithm \cite{williams1992simple} whose goal is to maximizes a reward, usually given in the form of a non-differentiable evaluation metric (here we use CIDEr since it is the main ranking metric used in the competition). In SCST, we first sample a caption and compute its CIDEr reward, then we compare it to the reward obtained by a test-time baseline (a greedy max decoding).
The sampled captions outperforming the baseline are given a positive weight, while under-performing samples are suppressed.

We perform SCST on CE models from randomly chosen epochs and a variety of initial random number generator seeds. We used a batch size of $80$ images/captions as well as the states learned in the ADAM optimizer during CE training.
To reduce the learning rate for SCST, we initialize the step number to be $50k$ plus the last step number of CE training.
Therefore, there is no warm up in SCST.
Finally, we perform SCST for a random number of additional epochs ranging from $15$ to $40$.\\

\noindent \textbf{Ensembling} The final step in our training pipeline is to form an ensemble from the obtained SCST models. At each step of caption generation, we first average the probabilities across the models for each vocabulary element and then sample the most likely token (greedy decoding). 

\subsection{Evaluation on EvalAI}
The evaluation of VizWiz competition is hosted on \textit{www.eval.ai}, which is an open source platform for evaluating and comparing various ML and AI algorithms at scale.
For this challenge, the participating teams submitted generated captions on the full VizWiz test dataset, consisting of 8,000 images.
This test dataset was used to support the two main evaluation phases: \textbf{test-dev}, consisting of 4,000 test images where teams could submit at most ten times per day, and \textbf{test-challenge}, containing all 8,000 test images where teams could submit at most five times over the course of the challenge.
Results from the test-challenge determined the challenge winners.
For our experimental evaluations, we report the results on the \textbf{test-dev} split, while the final competition results are shown on the \textbf{test-challenge} set.

\subsection{Post Processing of Captions}\label{sec:postprocessing}

In order to generate final caption submissions, we incorporate various post-processing steps.
Specifically, as shown in Figure \ref{fig_mmcap}, we apply three strategies: (a) De-tokenization, (b) self-BERT, and (c) OCR Maximization. While (a) is used to merge the sequences, generated from the decoder, to their untokenized representations in order to make it look more plausible and human-like, (b) and (c) are strategies that rank candidate captions.
For each image, we consider five candidate captions generated from multiple systems.
These systems are various ensemble models which achieved high CIDEr on the development set.
Note that steps (b) and (c) are used only for competition-purposes, since we observe that these steps help achieve better CIDEr on the validation set.
For self-BERT, we follow a similar methodology to that of self-BLEU \cite{zhu2018texygen} but we use BERT-score \cite{zhang2019bertscore} for better semantic similarity.
The goal of self-BERT is to select a caption from candidates which is most semantically similar to all the other candidates, and thus, the higher the self-BERT score, the better.
Similarly, for the OCR maximization step, we look at the coverage of tokens in generated image caption and tokens detected by the OCR modality. We notice that selecting a caption whose tokens has most overlap with the OCR tokens helps boost the CIDEr score.

\subsection{Ablation studies}
\label{sec:ablation}
In this section, we present our ablation studies to better understand the effect of each modality (IMG, OCR or OBJ) and the benefit of copy mechanism on the generated captions.
We build the models following the same CE+SCST pipeline as described above, i.e., the base models are first pre-trained with CE loss followed by SCST fine-tuning. For each selected study, we vary a few model hyper-parameters, such as the number of Transformer layers, hidden dimension or the initialization seed, to finally obtain the ensemble of multiple models, which is then evaluated on the test-dev evaluation dataset.
Additionally, we also show the results with and without the post-processing (Section \ref{sec:postprocessing}) to examine the additional improvement that this technique has on the captions. 

In Table \ref{tbl:ablation}, we study the effect of the data modality, which reveals that while incorporating each of the individual detectors provides gains in automatic scoring metrics above a baseline model that only considers image features, more substantial gains come from the OCR module.
This finding is in line with manual inspection of the data that highlighted the prevalence of captions that include a reference to text found in the image, as noted in Section \ref{subsubsec:ocr}. Also, the results in Table \ref{tbl:ablation} show that the post-processing provides an additional boost in the metrics (e.g., for CIDEr it is usually 2-3 points), which was quite valuable in this competition. 

In Table \ref{tbl:vizwiz_comp} (top 5 rows) we study the benefit of the copy mechanism (see Section \ref{sec:copy}). The setup is the same as before, except that  we use all the three data modalities (IMG, OCR and OBJ) and present the post-processed results. As can be seen, for a single model, there is a clear benefit of using the copy mechanism (compare lines 1 and 3 in the table). On the other hand, once the ensemble of such models is used, the advantage of copy-based models disappears (lines 2 and 4). However, combining all the copy and non-copy models together (ensemble of 90 models), which is our winning entry in the VizWiz competition, provides improvement across all metrics, especially CIDEr.

\begin{table}
\caption{Ablation studies. Each ensemble consists of 20 models (without copy mechanism) with varying number of layers, hidden dimension, and random seed. Each cell shows the value of a metric before and after post-processing the caption output. All the results are computed on the test-dev evaluation split.}
\resizebox{\textwidth}{!}{\begin{tabular}{lccccc}
 & BLEU4 & METEOR & ROUGE-L & CIDEr & SPICE \\ \midrule
IMG only & 25.11 / 25.97 & 21.01 / 21.22 & 48.22 / 48.55 & 66.8 / 69.41 & 15.83 / 16.04 \\ 
IMG+OCR & \textbf{25.88} / \textbf{26.83} & \textbf{21.49} / \textbf{21.73} & \textbf{49.17} / \textbf{49.54} & \textbf{72.45} / \textbf{75.51} & \textbf{16.23} / \textbf{16.47} \\ 
IMG+OBJ & 25.39 / 26.21 & 21.02 / 21.21 & 48.33 / 48.63 & 66.97 / 69.43 & 15.91 / 16.07
\end{tabular}}
\label{tbl:ablation}
\end{table}

\begin{table*}[!htb]
\centering
\caption{Evaluation results of different sets of models that are based on all three modalities (Image, OCR and OBJ), as well as use (and omissions) of the copy mechanism. Each cell shows the value of a metric after post-processing the caption output. The top 5 rows are evaluated on test-dev split, while the bottom row is the evaluation on the test-challenge set.}
\resizebox{\textwidth}{!}{\begin{tabular}{cccccccc}
\multicolumn{1}{l}{Evaluation} & \multicolumn{1}{l}{Copy Mechanism} & \multicolumn{1}{l}{Model Set} & \multicolumn{1}{l}{BLEU4} & \multicolumn{1}{l}{METEOR} &\multicolumn{1}{l}{ROUGE-L} & \multicolumn{1}{l}{CIDEr} & \multicolumn{1}{l}{SPICE} \\ \midrule
\multirow{5}{*}{test-dev} & \multicolumn{1}{c}{\multirow{2}{*}{Copy}} & Single & 23.66 & 20.34 & 47.34 & 64.48 & 14.81 \\ 
&\multicolumn{1}{c}{} & Ensemble (40) & 25.50 & 21.44 & 49.19 & 72.57 & 16.28 \\ 
\cmidrule{2-8}
&\multicolumn{1}{c}{\multirow{2}{*}{Non Copy}} & Single & 22.10 & 19.71 & 46.36 & 60.90 & 14.70 \\ 
&\multicolumn{1}{c}{} & Ensemble (40) & 27.43 & 22.20 & 50.10 & 78.80 & \textbf{17.35} \\ 
\cmidrule{2-8}
&\multicolumn{1}{c}{Copy + Non Copy} & Ensemble (90) & 27.31 & 22.13 & 50.02 & 80.38 & 17.10\\ \midrule  
test-challenge &\multicolumn{1}{c}{Copy + Non Copy} & Ensemble (90) & \textbf{27.44} & \textbf{22.25} & \textbf{50.20} & \textbf{81.04} & 17.00
\end{tabular}}
\label{tbl:vizwiz_comp}
\end{table*}

\begin{table*}[!htb]
\centering
\caption{CIDEr performance of the competition models based on the presence of text in an image. All values are taken from \protect \cite{slides_yinan} . Our captioner has a clear advantage when text is present confirming the benefits of using the OCR data modality. For images without text, we are slightly behind the SRC-B captioner.}
\begin{tabular}{ccccccc}
\multicolumn{1}{l}{} & Ours & SRC-B & aburns & LittlePanda & iim & Baseline \\ \midrule
with text & \textbf{91.60} & 77.78 & 67.88 & 64.76 & 64.17 & 64.81 \\ 
no text & 59.80 & \textbf{60.43} & 54.90 & 50.37 & 49.7 & 49.01
\end{tabular}%
\label{tbl:textnotext}
\end{table*}

\begin{table*}[!htb]
\centering
\caption{CIDEr performance of the competition models based on the image quality. All values are taken from \protect\cite{slides_yinan}. The definition of easy, medium, and hard is given the text. In all three categories, our captioner shows a clear advantage over the competing approaches.}
\begin{tabular}{ccccccc}
\multicolumn{1}{l}{} & Ours & SRC-B & aburns & LittlePanda & iim & Baseline \\ \midrule
Easy & \textbf{88.76} & 77.48 & 68.15 & 64.74 & 64.07 & 64.52 \\ 
Medium & \textbf{66.71} & 63.26 & 56.67 & 51.94 & 51.43 & 51.27\\ 
Hard & \textbf{41.31} & 39.87 & 35.30 & 33.83 & 33.39 & 31.84
\end{tabular}%
\label{tbl:easyhard}
\end{table*}

\subsection{Competition Results}

As discussed in Section \ref{sec:ablation}, our best captioner is an ensemble of 90 models, employing all three data modalities (IMG, OCR and OBJ) and a mix of copy and non-copy mechanism models, followed by post-processing of the generated sentences.
Evaluating this model on the test-challenge split gives the result presented at the bottom of Table \ref{tbl:vizwiz_comp}.
These test-challenge scores represent the largest values we obtained across most of the metrics, achieving the highest CIDEr score of 81.04.

Tables \ref{tbl:textnotext} and \ref{tbl:easyhard} (all the evaluation numbers are taken from \cite{slides_yinan}) compare our method to a few other challenge participants in terms of CIDEr score performance on the test-challenge evaluation set. In what follows, we first briefly review some of these competitor models and then discuss the table results. 

\noindent \textbf{Baseline} \cite{huang2019attention}: This is the original baseline approach used by the organizers.
The captioner is based on Attention on Attention module, which refines the traditional attention mechanism to better determine the relevance between the query and the attended features. 

\noindent \textbf{SRC-B} \cite{video_srcb} [2nd place]: This is a captioner based on X-Linear Attention Networks \cite{pan2020x} which uses Faster-RCNN image features pretrained on Visual Genome \cite{anderson2018bottom}. The model is trained using the Bayesian SCST-based approach \cite{bujimalla2020b}.

\noindent \textbf{aburns} \cite{video_aburns} [3rd place]: The captioner is an extension of AoANet \cite{huang2019attention} where the visual features from original attention mechanism are fed into another attention module over the OCR semantic features. These OCR tokens are first extracted using GCloud OCR Vision Cloud API and then converted into fastText feature representation.

Table \ref{tbl:textnotext} shows the comparison of the models' performance on images with and without the text. Our captioner is a clear top performer when text is present confirming the benefits of using the OCR data modality. On the other hand, for images without the text, we are slightly behind the \textbf{SRC-B} captioner.

Finally, Table \ref{tbl:easyhard} compares the models based on the image quality. 
The organizers defined hard images to be 25\% of all images that have a maximum CIDEr score of less than 50 across all the challenge submissions. 
The easy images are those that have a minimum CIDEr of 100 across all submissions, which account for 6.73\% of all the image. 
The rest of them are deemed medium difficulty. 
Across all three categories of difficulty, our captioner shows a clear advantage over the competing approaches.


\begin{figure*}[!htb]

 	\centering
 	\begin{subfloat}
 		\centering
 		\resizebox{0.99\linewidth}{!}{
 			\begin{TAB}(c){|c|c|}{|c|c|c|c|}
 				\includegraphics[width=1.5in, height=1.5in, keepaspectratio]{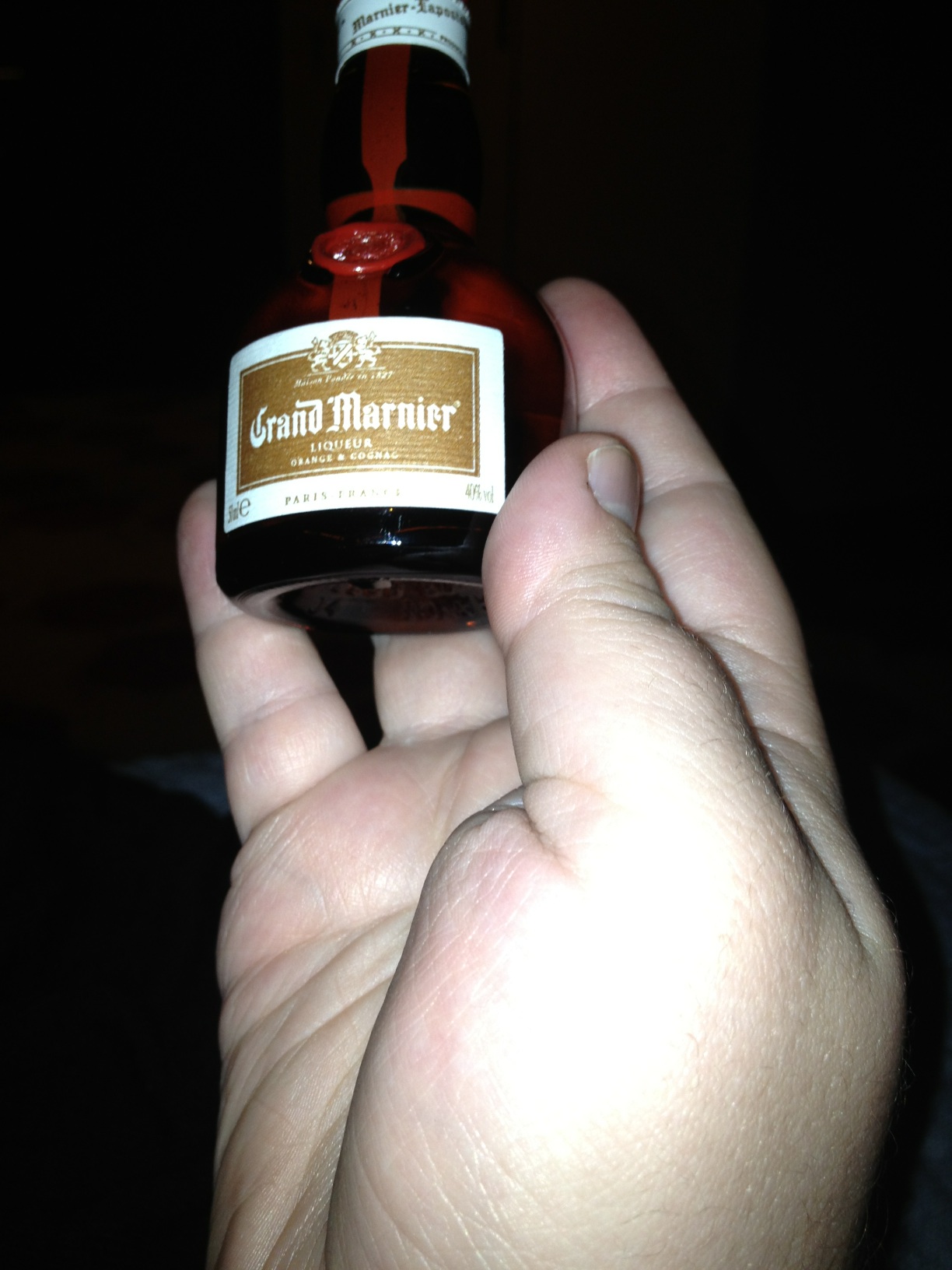} &
 				\raisebox{0.0cm}{\includegraphics[width=1.5in, height=1.5in, keepaspectratio]{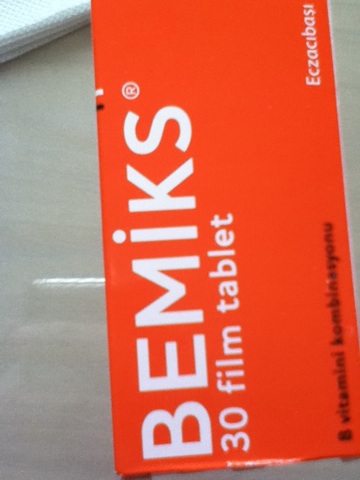}} \\
 				\parbox[][][c]{3.5in}{%
 					\texttt{V}: a person is holding a bottle of cognac\\
 					\texttt{C}: a bottle of champagne\\
 					\texttt{G}: a bottle of red wine\\
 					\texttt{OCR}: cognac, liqueur, marnier, trand\\				
 					\texttt{Obj}: bottle, person
 				}
 				&
 				\parbox[][][c]{3.5in}{%
 					\texttt{V}: a red box of bemiks sitting on a white surface\\
 					\texttt{C}: a sign for a parking lot\\
 					\texttt{G}: a sign for a public parking lot\\
 					\texttt{OCR}: bemiks, film, tablet\\
 					\texttt{Obj}: book
 				}
 				\\
 				\includegraphics[width=1.5in, height=1.5in, keepaspectratio]{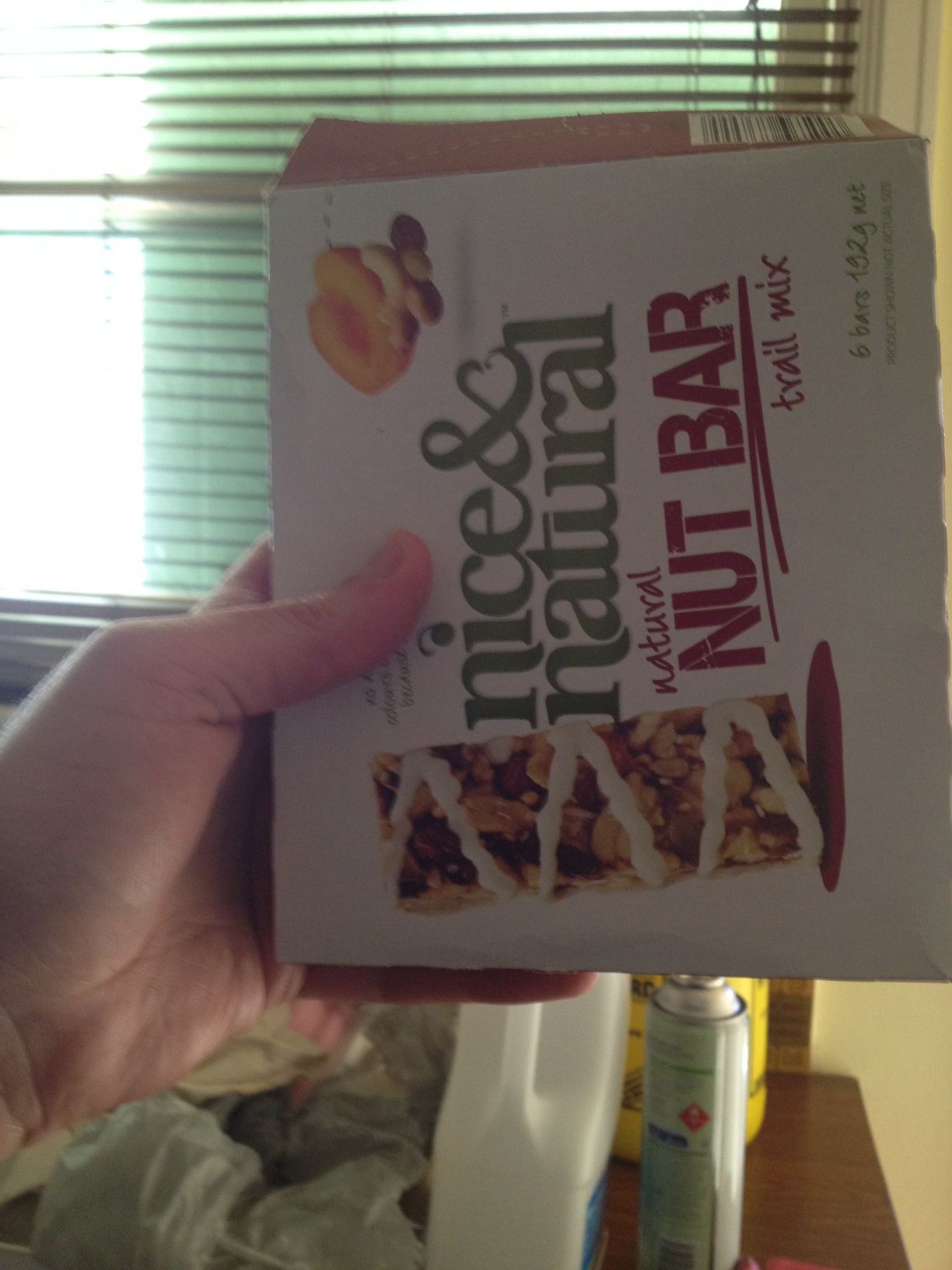} &
 				\raisebox{0.0cm}{\includegraphics[width=1.5in, height=1.5in, keepaspectratio]{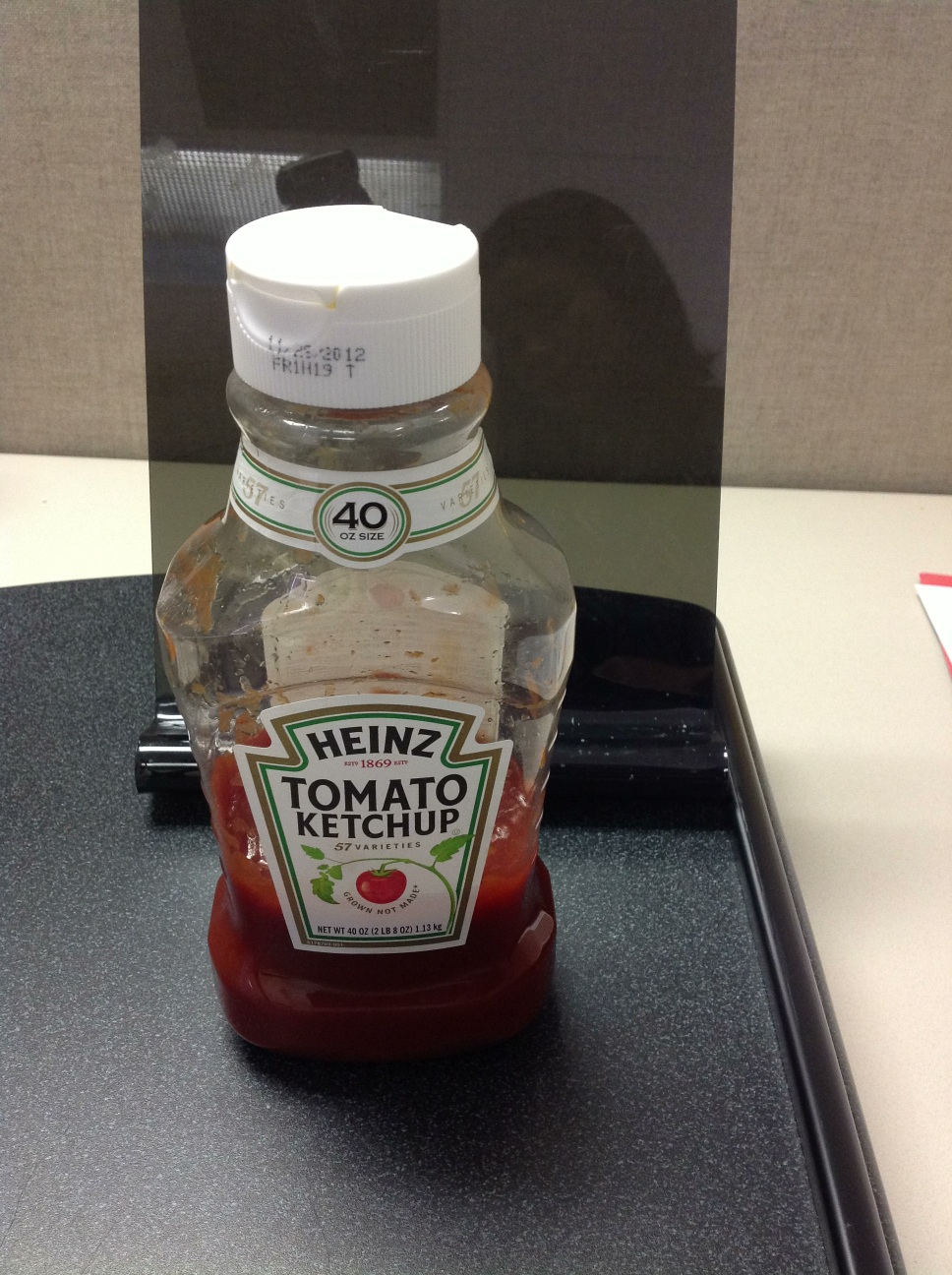}} \\
 				\parbox[][][c]{3.5in}{%
 					\texttt{V}: a person is holding a box of trail mix\\
 					\texttt{C}: a hand holding a piece of paper with a red and white background\\
 					\texttt{G}: a hand holding a piece of cake\\
 					\texttt{OCR}: actual, bars, net, product, shown, size, nut, bar, trail, mix\\
 					\texttt{Obj}: book, bottle, person, pizza
 				}
 				&
 				\parbox[][][c]{3.5in}{%
 						\texttt{V}: a bottle of heinz tomato ketchup sitting on a counter\\
 					\texttt{C}: a bottle of milk\\
 					\texttt{G}: a jar of brand\\
 					\texttt{OCR}: 2012, heinz, tomato, ketchup, ozsize\\
 					\texttt{Obj}: bottle, dining table
 				}
 			\end{TAB}%
 		}
 		\caption{Comparison of our model (\texttt{V}) against generic captioners (\texttt{C}, \texttt{G})  on various images from indoor scenes.}
 		 \label{fig:indoor_cap}

 	\end{subfloat}%
 \end{figure*}

 \begin{figure}[ht!]
\centering
\includegraphics[width=0.8\linewidth]{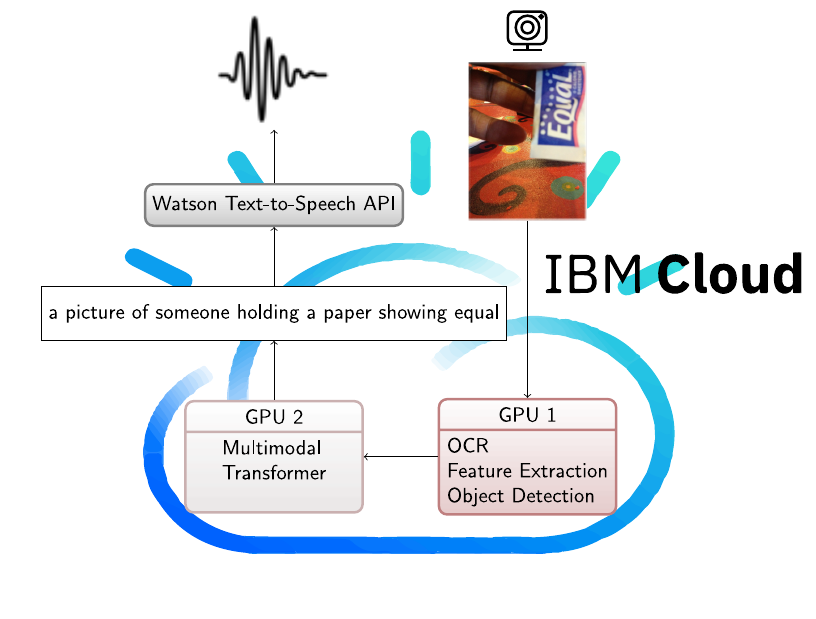}
\caption{Assistive captioner real time demo in the cloud.}
\label{fig_demo}
\end{figure}

\begin{figure*}[!htb]
 	\centering
 	\begin{subfloat}
 		\centering
 		\resizebox{0.99\linewidth}{!}{
 			\begin{TAB}(c){|c|c|}{|c|c|c|c|}
 				\includegraphics[width=1.5in, height=1.5in, keepaspectratio]{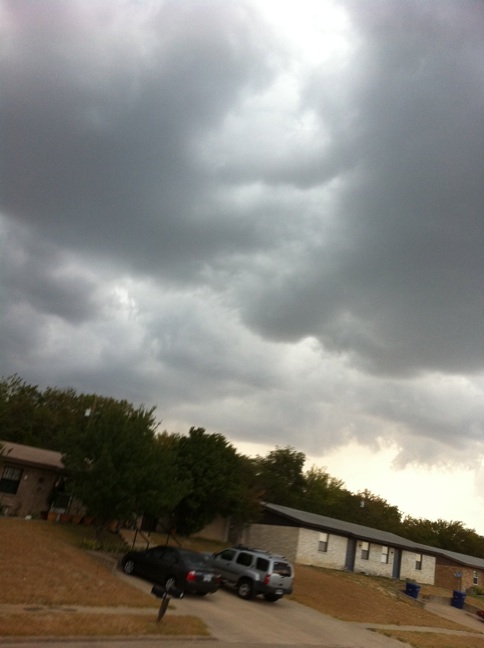} &
 				\raisebox{0.0cm}{\includegraphics[width=1.5in, height=1.5in, keepaspectratio]{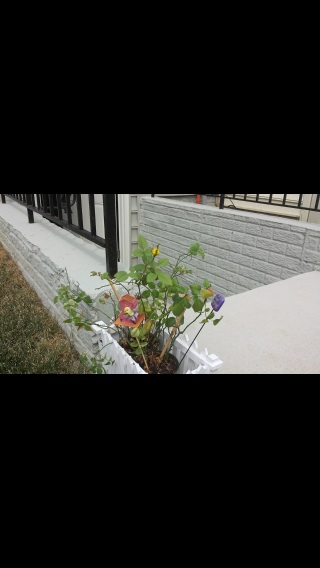}} \\
 				\parbox[][][c]{3.5in}{%
 					\texttt{V}: a cloudy sky with a blue sky and cars in the\\
 					\texttt{C}: a view of the house\\
 					\texttt{G}: a storm over a city\\
 					\texttt{OCR}: -\\				
 					\texttt{Obj}: car
 				}
 				&
 				\parbox[][][c]{3.5in}{%
 				\texttt{V}: a porch with a white and green flowers in front of a house\\
 					\texttt{C}: a red flower pot on a brick wall\\
 					\texttt{G}: a beautiful red flower in the corner of a brick wall\\
 					\texttt{OCR}: -\\				
 					\texttt{Obj}: -
 				}
 				\\
 				\includegraphics[width=1.5in, height=1.5in, keepaspectratio]{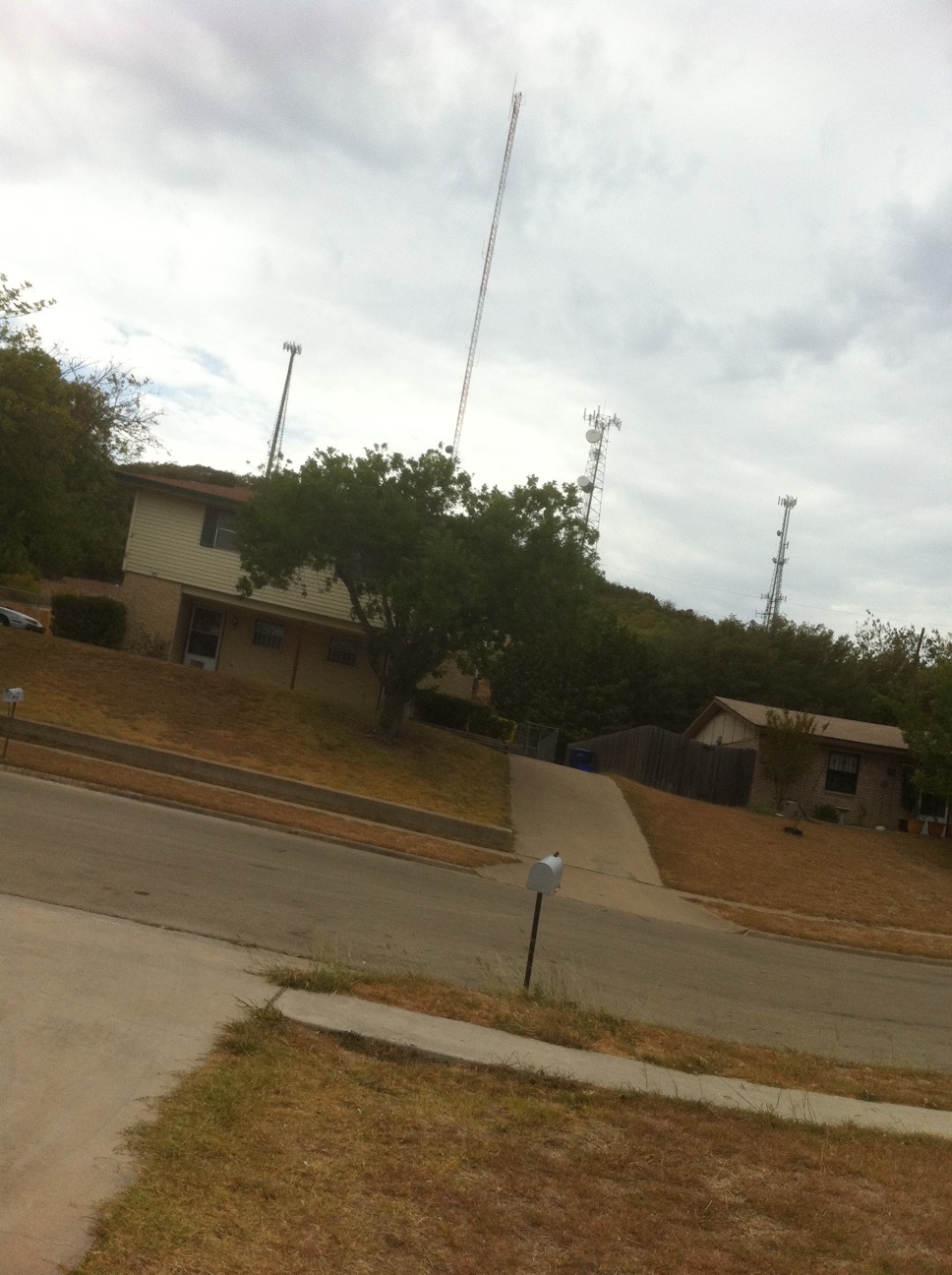} &
 				\raisebox{0.0cm}{\includegraphics[width=1.5in, height=1.5in, keepaspectratio]{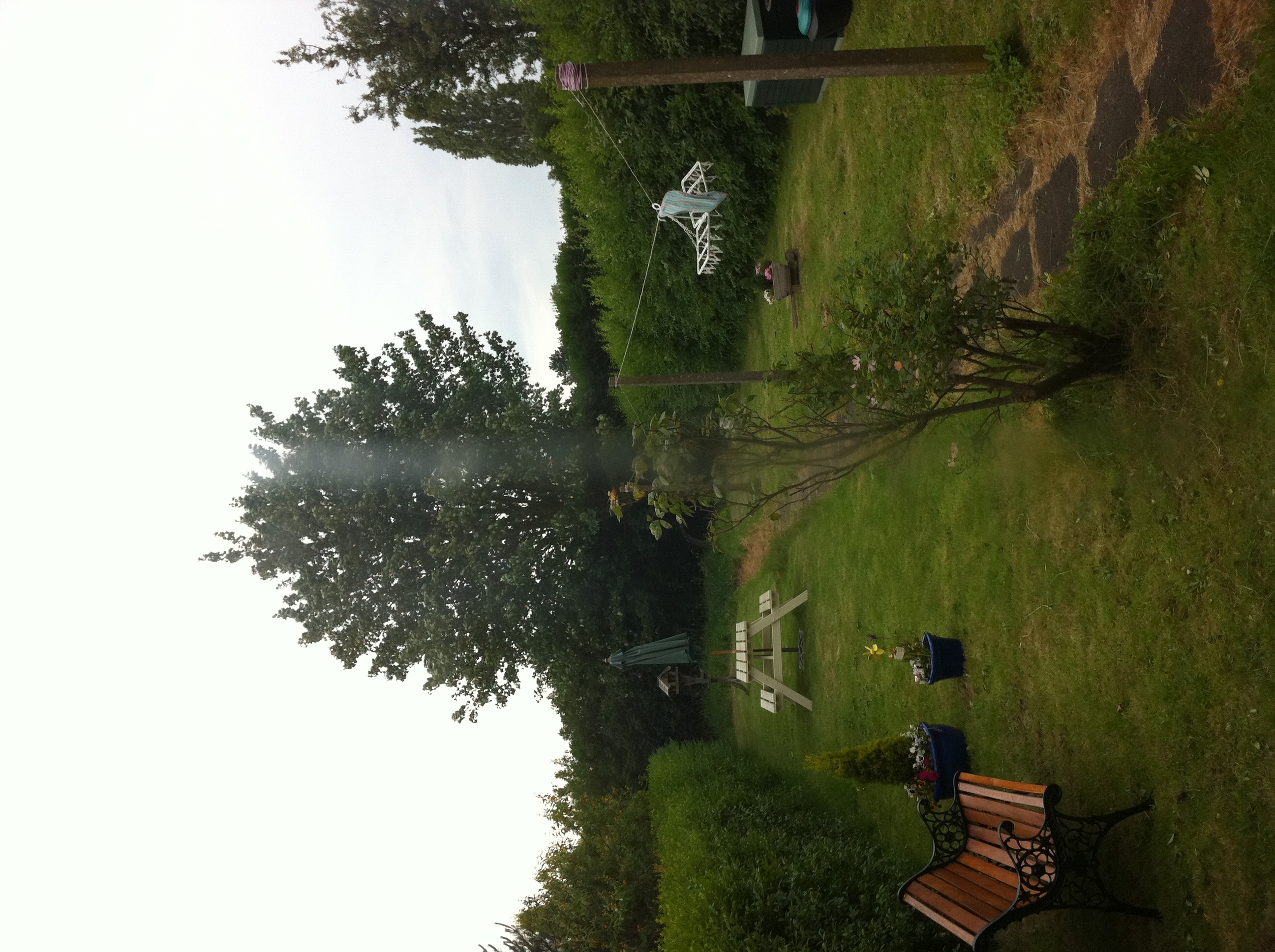}} \\
 				\parbox[][][c]{3.5in}{%
 					\texttt{V}: a street with a house and a car in front of a house\\
 					\texttt{C}: the house is on the side of the road\\
 					\texttt{G}: the road to the property\\
 					\texttt{OCR}: -\\				
 					\texttt{Obj}: car
 				}
 				&
 				\parbox[][][c]{3.5in}{%
 						\texttt{V}: a backyard with a green grass and trees in the on the\\
 					\texttt{C}: a picnic table in a garden\\
 					\texttt{G}: a picnic table in a garden\\
 					\texttt{OCR}: -\\				
 					\texttt{Obj}: bench
 				}
 			\end{TAB}%
 		}
 		\caption{Comparison of our model (\texttt{V}) against generic captioners (\texttt{C}, \texttt{G}) on various images from outdoor scenes.}
 		\label{fig:outdoor_cap}

 	\end{subfloat}%
 \end{figure*}

\section{Discussion  of Goal-Oriented Captioning versus Generic Captioning} \label{sec:qualitative}

In order to provide more nuanced qualitative findings, we look at captions generated by our system for two categorical scenes - indoor (Figure ~\ref{fig:indoor_cap}) and outdoor (Figure \ref{fig:outdoor_cap}). In both these figures, we select some of the images and the captions generated by our best ensemble system, \texttt{V}, along with the output from OCR and OBJ modalities.
For comparison, and, in the spirit of approach followed by \cite{vizwiz_cvpr}, we also show the outputs from models trained on a unified dataset combining MS-COCO and GCC \cite{sharma2018conceptual} data, with only image modality. \texttt{C} is the model selected with highest CIDEr on MS-COCO development dataset whereas, \texttt{G} is the model selected with highest CIDEr on GCC development of the dataset.
Note that \texttt{C} and \texttt{G} models are trained on image modality alone and serve as generic image captioning systems, whereas \texttt{V} is trained with all three modalities leading to more meaningful and informative captions, and thus, making it a better task-/ goal-oriented system.

For the images provided in Figure \ref{fig:indoor_cap}, it can be noticed that the role of OCR modality is crucial for generating the captions. A particularly interesting case is of the caption containing the token \textit{'bemiks'} (top right image in Figure \ref{fig:indoor_cap}). The token \textit{'bemiks'} does not appear in the training image captions vocabulary and our model manages to use this token, detected by OCR modality, during the decoding step.
This underscores the importance of copy mechanism on OCR modality for the scene. 
The role of OCR modality can further be supported in the ketchup bottle example (bottom right image in Figure \ref{fig:indoor_cap}), where the OCR tokens \textit{'heinz'}, \textit{'tomato'}, \textit{'ketchup'} are captured by the system.
Captions in Figure \ref{fig:outdoor_cap} also highlight the positive impact of the OBJ modality in our system. For example, in the bottom left image in Figure~\ref{fig:outdoor_cap}, our system manages to focus on the tokens generated by the OBJ modality while the \texttt{C} and \texttt{G} models fail to mention anything about the car present in the image.
This is also an example which shows the more fine-grained capabilities of the OBJ modality, which detects a \textit{'car'} even though it is barely visible in the scene and quite easy to miss.

\section{Image Captioning As an Assistive Technology : Real-Time Demo }\label{sec:demo}

Our real-time assistive image captioning demo summarized in Figure \ref{fig_demo} is designed as a progressive web application (PWA) with the intent to be used in a cross-platform as well as cross-device manner. Using a PWA allows the visually impaired to take as input an image captured via either a cellphone camera, webcam or uploaded locally from a variety of systems. 
After the image has been obtained, it is sent to the first GPU in the pipeline. 
Three different transforms of the image are performed. One transformation consists of feature extraction via the ResNeXt network.
The next transformation is an extraction of objects defined by the object detector.
Lastly, text found in the image is read from the image and extracted by the OCR module. These separate features are combined into single multimodal feature and then sent to the second GPU. This multimodal feature is then pushed into the Transformer and the captions are generated in the form of a string of text using greedy max decoding. 
Once completed, the generated caption string, the detected objects and texts from OCR are sent through the Watson Text-to-Speech API to be transformed into an mp3 file on a cloud server. After the mp3 file is returned, we play the audio file for the user. 
The audio file consists of the caption, listing of detected texts and objects. 
We also visualize the image, caption, objects and the words detected by the OCR to the screen (see the video of the real time demo on \url{https://github.com/IBM/IBM_VizWiz} ). 
The realization of this pipeline into a real time demo was made possible using flask \cite{grinberg2018flask}, a Python web package.

\section{Conclusion}

In this paper, we presented our contributions to realizing the potential of image captioning as an assistive technology for the visually impaired, focusing on our winning submission to the VizWiz 2020 Image Captioning Challenge.
We described the multiple set of functionalities needed by a machine learning system to achieve meaningful task-oriented captioning, including object and text detection and recognition.
Dictionary-guided rotation-invariant text recognition is a crucial component of our system, as images captured by the visually impaired may have severe quality issues in terms of orientation, blur, and occlusion. Our winning entry to the 2020 VizWiz Grand Challenge and its real time demo is a step towards translating the success of deep multimodal models on curated datasets, such as MS-COCO, to useful assistive technology. A more interactive system, in the form of visual dialog, would be an important next step, as it would allow for rounds of feedback between the visually impaired users and the system in order to facilitate users in reaching their goals and completing tasks at hand. This type of dialog will help with occlusions and bad image conditions that the system would eventually detect, e.g.\ by asking for additional images from the user, and would refine the system response in case a user was not satisfied with a provided caption.
Applications of computer vision in accessibility and science for social good will require this feedback loop as well as the goal-oriented components we highlighted throughout this work.



%



  \section*{Acknowledgment}
The authors would like to thank the support of the science for social good program in IBM Research AI, and the MIT-IBM Watson AI lab for providing cloud machines for the real time demo. Authors are also thankful to the 2020 VizWiz Grand Challenge competition organizers, in particular Danna Gurari and  Yinan Zhao.

\vskip 0.2in
\bibliographystyle{apacite}
\bibliography{pami_jair}

\end{document}